\documentclass[letterpaper, 10 pt, conference]{ieeeconf}  

\IEEEoverridecommandlockouts                              

\usepackage{graphicx}
\usepackage{algorithm}
\usepackage{array}
\usepackage{amsmath}
\usepackage{url}
\usepackage{enumerate}
\usepackage{multicol}
\usepackage{textcomp}
\usepackage{multirow}
\usepackage{cite}
\usepackage{fmtcount}
\usepackage{diagbox}
\usepackage{outlines}

\usepackage{color}
\definecolor{Change}{rgb}{0.5,0.1,0.1}

\usepackage{threeparttable}

\usepackage{changes}
\definechangesauthor[name={Per cusse}, color=orange]{per}

\usepackage[noend]{algpseudocode}
\usepackage{algorithm}
\usepackage{algorithmicx}
\usepackage{varwidth}

\usepackage{amsfonts}

\usepackage{tikz}

\title{\LARGE \bf
Scene Context Based Semantic Segmentation for 3D LiDAR Data in Dynamic Scene
}

\author{Jilin~Mei,~\IEEEmembership{Member,~IEEE,}
	and Huijing~Zhao, ~\IEEEmembership{Member,~IEEE,}
	\thanks{J. Mei and H. Zhao are with the Peking University, with
		the Key Laboratory of Machine Perception (MOE), and also with the School of
		Electronics Engineering and Computer Science}
	\thanks{Correspondence: H. Zhao, zhaohj@cis.pku.edu.cn.}
}

\begin{document}

\maketitle
\thispagestyle{empty}
\pagestyle{empty}

\begin{abstract}
	We propose a graph neural network(GNN) based  method to incorporate scene context for the semantic segmentation of 3D LiDAR data.  The problem is defined as building a graph to represent the topology of a center segment with its neighborhoods, then inferring the segment label. The node of graph is generated from the segment on range image, which is suitable for both sparse and dense point cloud. Edge weights that evaluate the correlations of center node and its neighborhoods are automatically encoded by a neural network, therefore the number of neighbor nodes is no longer a sensitive parameter. A system consists of segment generation, graph building, edge weight estimation, node updating, and node prediction is designed. Quantitative evaluation on a dataset of dynamic scene shows that our method has better performance than unary CNN with 8\% improvement, as well as normal GNN with 17\% improvement.
\end{abstract}

\section{Introduction}

\begin{figure}
	\centering
	\includegraphics[width=0.47\textwidth]{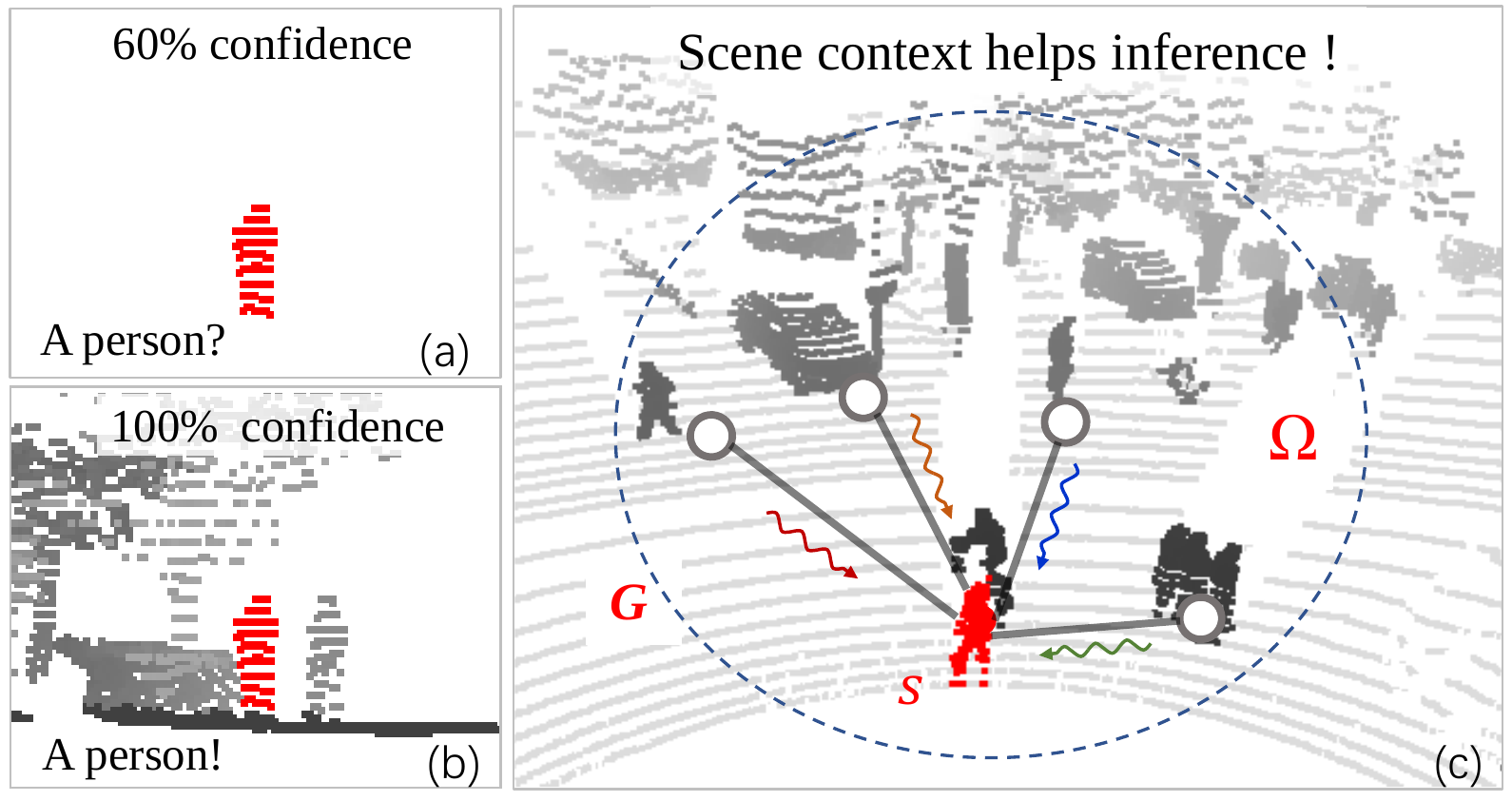}
	\caption{Illustration of scene context based semantic segmentation. (a) The red segment is foreground, and it is not confident deciding the segment is a person. (b) Under the help of context information, we find the person is standing nearby trees with another one. (c) The semantic segmentation is defined as representing the topology of a center segment $s$ with its neighborhood $\Omega$ using a graph $G$, 
		and inferring the label of $s$ by a GNN-based classification.	
	}
	\label{fig:intro1}
\end{figure}

LiDAR sensor has been applied in autonomous driving systems since last decade \cite{urmson2008autonomous}. Compared with detection\cite{Shi2018PointRCNN3O}, semantic segmentation of 3D LiDAR data supports more sophisticated sensing capability which is an important part to build smarter and safer systems.
Semantic segmentation can be defined as the process of finding semantic labels for each point/segment, that is, classifying each data unit. In one data frame, one data unit could be the foreground as the red segment shown in Fig. \ref{fig:intro1}, and it also could be the context of other data units at the same time.

In our previous work \cite{Mei2019SupervisedLF}, although the context is not modeled explicitly, the classification accuracy is largely improved when the sample includes context. On the other hand, the researches on scene graph\cite{Zellers2017NeuralMS,Yang2018GraphRF} also depict the effectiveness of scene context. All of these motivate us to find a way to model the relationship between foreground and context. As shown in Fig. \ref{fig:intro1}, the connections between foreground and context are captured by a graph model, then the semantic segmentation is converted as gathering information from context and determining the label of foreground. We believe the explicit context information will help the classification of foreground, and it also assists high-level object relationship reasoning, for example, scene graph generation\cite{Li2017SceneGG}.   

The graph neural network(GNN)\cite{Zhou2018GraphNN} provides a basic solution for our task. As shown in Fig. \ref{fig:intro1}, the foreground and context are abstracted as nodes, and their relationships are represented as edges where nodes messages pass through. In GNN, node/edge representation and message passing are all captured by deep learning methods. The GNN about scene context could be divided into two categories.

The first one is scene graph generation on 2D
camera image. \cite{Zellers2017NeuralMS,Li2017SceneGG} directly retrieve pairwise relationships among instances from image, which is defined as scene graph generation. Our task is very similar with scene graph, however, we can not directly apply it to 3D LiDAR data. The main reason lies in that the training steps of scene graph are highly dependent on dataset, for example, both instance and relationship annotations are required. To the best of our knowledge, available 3D LiDAR point cloud datasets\cite{Hackel2017Semantic3DnetAN,behley2019iccv} can not satisfy the training steps of scene graph at all.

The second one is semantic segmentation on 3D point cloud. Based on the node type, these methods could be separated into two streams: point-wise graph and cluster-wise graph. In the first stream\cite{Qi20173DGN,Wang2019GraphAC}, each point represents one node, which causes up to 80\% of the time is wasted on structuring the sparse data\cite{Liu2019PointVoxelCF}. In the second stream, \cite{Landrieu2018LargeScalePC} builds the node on point cluster(named supperpoint), while the edge feature is handcrafted; the shortcoming is that we may design new edge features for new datasets. These methods are difficult directly applying to sparse 3D LiDAR data collected by autonomous driving platforms. Our method belongs the second stream,  but a special GNN-based classifier, that considers both the local compactness of node and the generality of edge feature, is proposed to incorporate scene context in the semantic segmentation of 3D LiDAR data. 

This research proposes a scene context based semantic segmentation of 3D LiDAR data in urban dynamic scenes, where a graph is built to represent the topology of a center segment with its neighborhoods, and the segment label is inferred by a GNN-based classifier. The major advantages of our method are: 1) Graphs are built with nodes representing data segments, which are extracted by examines consistency of 3D points in range frame and is suitable for both sparse and dense point cloud; 2) Edge weights are automatically encoded by a neural network, and the predefined edge feature\cite{Landrieu2018LargeScalePC} is not required; 3) The number of neighbor nodes is no longer a sensitive parameter, as their correlations with the center node are evaluated by edge weights and estimated during graph updating. Experiments are conducted on a dataset of dynamic scene \cite{Pan2019SemanticPOSS}, and quantitative evaluation shows that the proposed GNN classifier has better performance than unary CNN method with 8\% improvement, as well as normal GNN method with 17\% improvement.

The remainder of this paper is structured as follows. Sect. \uppercase\expandafter{\romannumeral2} discusses related work. The proposed method is presented in Sect. \uppercase\expandafter{\romannumeral3}. Sect. \uppercase\expandafter{\romannumeral4} shows the implementation details. Sect. \uppercase\expandafter{\romannumeral5} presents the experimental results. Finally, we draw conclusions in Sect. \uppercase\expandafter{\romannumeral6}.

\section{Related Works}

LiDAR-based semantic segmentation has been studied since last decade. We firstly review semantic segmentation methods, then discuss graph neural network applied to semantic segmentation.

\subsection{Semantic Segmentation of 3D LiDAR Data}
Semantic segmentation goes through two periods: traditional method and deep learning method. In traditional method\cite{munoz2009onboard,zhao2010scene,lu2012simplified}, semantic segmentation is the process of classifying each data unit. Some methods directly classify each laser point. Others\cite{zhao2010scene} firstly divide data frame into geometric consistent segments, and then each segment will be classified. No matter which way of data unit definition, the traditional method designs handcrafted features for each element. Some methods\cite{lu2012simplified} consider the spatial relationships between elements via graph model, such as Markov random field or conditional random field; Each data unit constitutes a node, and the geometric relationship between nodes constitutes an edge. The disadvantage of traditional method is the need for expert experience to adjust features, however, the development of traditional methods constructs the basic framework of semantic segmentation.

With the success of deep learning in image semantic segmentation\cite{GarciaGarcia2017ARO}, it is also applied to point cloud data. A direct way is projecting 3D point cloud into 2D image, e.g., range image or top-view image, so that we can use the popular networks in image segmentation; In \cite{Caltagirone2017FastLR}, the point clouds are encoded by top-view images and a simple fully convolutional neural network (FCN) is used. Raw LiDAR data is sparse and unorder. Another stream adopts 3D representations, e.g., voxel occupancy to make LiDAR data grid-aligned\cite{hackel2017isprs}. The voxel representation is further improved in OctNet\cite{riegler2017octnet}. PointNet\cite{qi2017pointnet} directly takes raw point clouds as input, and a novel type of neural network is designed with multilayer perceptrons (MLPs). Pointnet is widely used in feature extraction of point cloud in the following research. These deep learning methods either assume that each data unit is independent or only consider the context information in a small area. However, elements in one scene have strong correlations, e.g., people tend
to walk nearby trees, as can be seen in Fig. \ref{fig:intro1}(b).  

\subsection{Graph Neural Network Applied to Semantic Segmentation}
Graph neural network\cite{Zhou2018GraphNN} is an effective method to extract long-range contextual information. Our task is highly relevant to scene graph generation\cite{Li2017SceneGG} which includes two stages: firstly detect instances and then predict the relationships among instances. By contrast, our task firstly determines the relationships between the node and its neighbors, then predict the label of node. Scene graph generation is discussed on camera image. and the question is that could we directly apply it when 3D point cloud is converted into 2D format?  The obstacle lies in that the training steps of scene graph are highly dependent on dataset, for example, both instance and relationship annotations are required. To the best of our knowledge, available 3D LiDAR point cloud datasets can not satisfy the training steps of scene graph at all.

Graph neural network is also discussed on point cloud semantic segmentation. Based on the node type, these methods could be separated into two streams: point-wise graph \cite{Qi20173DGN,Wang2019GraphAC}and cluster-wise graph\cite{Landrieu2018LargeScalePC}. In GAC\cite{Wang2019GraphAC}, each point builds one node and the edge feature is automatically determined via attention layers. GAC can be seen as an extension of pointnet++\cite{Qi2017PointNetDH}. The shortcoming of this point-based network is that up to 80\% of the time is wasted on structuring the sparse data which have rather poor memory locality, not on the actual feature extraction\cite{Liu2019PointVoxelCF}.  As for the cluster-wise graph, \cite{Landrieu2018LargeScalePC} proposes supperpoint graph. Each superpoint is a cluster of the raw point cloud, whereas the edge features are predefined which may cause new features are designed for new datasets. Inspired by \cite{Landrieu2018LargeScalePC}, we propose a special GNN based classifier, that considers both the local compactness of node and the generality of edge feature, to incorporate scene context in semantic segmentation of 3D LiDAR data. 

\section{METHODOLOGY}
\label{sect:method}

\begin{figure*}
	\centering
	\includegraphics[width=0.9\textwidth]{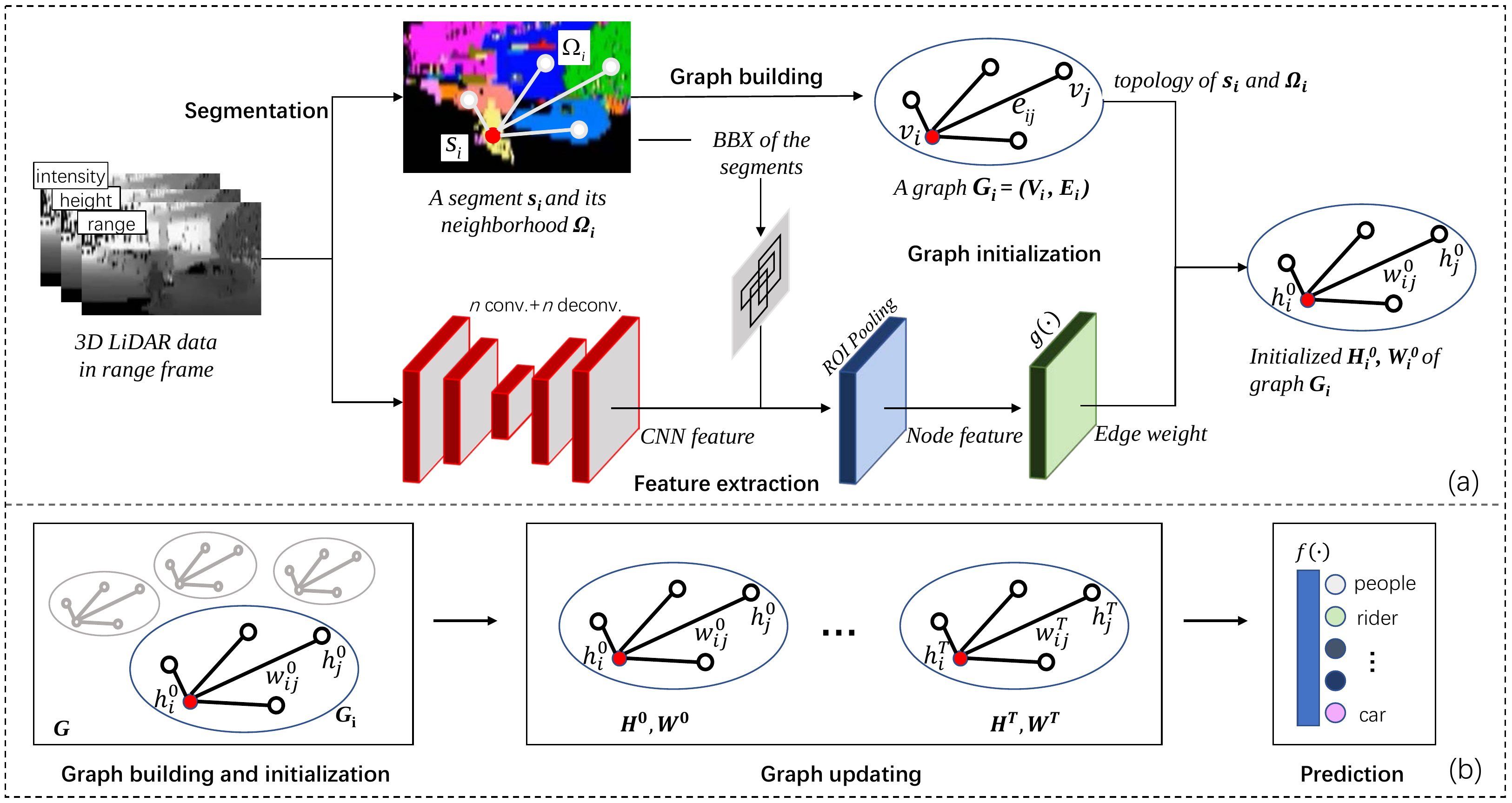}
	\caption{The framework of scene context based semantic segmentation. (a) Graph building and initialization. (b) graph updating and node prediction.  }
	\label{fig:framework}
\end{figure*}

\subsection{Problem Definition} 
Let $s$ denotes a segment of 3D LiDAR data, which is extracted by a segmentation method, e.g. a region growing one, that examines consistency of 3D points in range frame.
Let $\Omega$ be a set of $M$ nearest segments of $s$.
The problem in this work is to map $s$ to a label $x \in \{1,...,K\}$ by inferring with its neighborhood $\Omega$.
\begin{equation}
F_{\theta} : s | \Omega \to x \in \{1,...,K\} 
\end{equation}

In this research, a graph neural network\cite{Zhou2018GraphNN} based classifier is developed as illustrated in Fig.\ref{fig:framework}.
Given a data sample of a center segment $s_i$ and its neighborhood $\Omega_i$, an undirected graph $G_i=\{V_i,E_i\}$ is first built to represent the topology, where $V_i=\{v_i\}$ is a set of nodes corresponding each segment, and $E_i=\{e_{ij}\}$ are edges linking $v_i$, the center node, with all neighbor ones $v_j, j\in \Omega_i, j \neq i$.

Let $H_i=\{h_i\}$ and $W_i=\{w_{ij}\}$ be the node and edge states of $G_i$ respectively. 
$G_i$ is then initialized to find $H_i^0=\{h_i^0\}$ and $W_i^0=\{w_{ij}^0\}$ that can be formulated as below.
\begin{eqnarray}
&&\Theta(v_i) \to h_i^0    \\
&&g(\Theta(v_i), \Theta(v_j)) \to w_{ij}^0
\end{eqnarray}
In this research, $h_j^0$ is a feature vector that is initialized through a feature extraction procedure $\Theta(\cdot)$. $w_{ij}^0$ is a ratio/weight describing the relationship strength of nodes $v_i$ and $v_j$ that is calculated by an operator $g(\cdot)$.

The node and edge values are updated, and $H_i^T$ and $W_i^T$ are obtained after $T$ iteration steps. A semantic label of $v_i$ is finally predicted as below
\begin{equation}
\label{eq1}
f(h_i^T) \to \hat{x}_i\in \mathbb{R}^{1\times{K}}.
\end{equation}
$\hat{x}_i$ is the prediction label of $s_i$, and $f(\cdot)$ is a prediction operator.

Let $F_\theta$ denotes the above graph neural network-based classifier.
Given a set of supervised samples $X=\{s_i, \Omega_i, x_i\}_{i=1}^N$, where $x_i$ is a label of $s_i$ that is annotated by a human operator, the problem of learning $F_\theta$ can be formulated as finding the best $\theta^*$ that minimize a loss function $L$ as below.
\begin{equation}
\theta^{*}=\mathop{\arg\min}_{\theta}L(X; \theta)
\end{equation}

Below, we detail the edge weight estimations, the node updating, the node prediction and the loss function.

\subsection{Edge Weight Estimation}

If we assume each node $v_i$ is independent, \cite{Mei2019SupervisedLF} has validate this assumption is not conducive to the correct classification of $v_i$; on the contrary, the contextual information enhance the node representation. Therefore, we hope the messages of neighbor nodes could propagate to the target node. $v_i$ has multiple neighbors, obviously each neighbor should has different importance to $v_i$; for example, a walker gives more attentions to moving objects such as car and rider, whereas less attentions to static objects.  A weight/attention mechanism is designed to perceive contextual information\cite{Velickovic2017GraphAN}: $g_{\theta_{e}}: \mathbb{R}^{s}\times{\mathbb{R}^{s}} \to \mathbb{R}$. The function $g_{\theta_{e}}$ is shared by all edges in the graph model. The relationship of two adjacent nodes is defined as below: 
\begin{equation}
\label{lab_edge}
\tilde{e}_{ij} = g_{\theta_{e}}([h_i, h_j]),
\end{equation}
where $h_i$ and $h_j$ are the hidden states of $v_i$ and $v_j$; the operator $[\cdot]$ means vector concatenation. $g_{\theta_{e}}$ can be implemented by any differential function, the simplest one is a linear model, and we use a multiple layer perception(MLP) in this paper. The above equation depicts the importance of $v_j$ to $v_i$ and vice verse, as the graph is undirected. Obviously, we do not explicitly model the edge like designing an edge feature, but represent it from pair-wise nodes. The Eq.(\ref{lab_edge}) allows each node's message passing to the other node, and it can capture any long-range contextual information. In practice, we only consider limited neighbors of $v_i$, that is, the range of context is adjusted by $\Omega_i$. In this way, the edge weight is defined as:
\begin{equation}
w_{ij} = \frac{1}{|\Omega_i|}\frac{exp(\tilde{e}_{ij})}{\sum_{j\in\Omega_i}exp(\tilde{e}_{ij})}.
\end{equation}	
\textbf{Relationship to convolution filter.} The standard convolution operator also captures contextual information, but the perception field is small, and the size is defined by filter kernels whose sizes are 3x3 or 5x5 etc. The advantage of Eq.(\ref{lab_edge}) is that the perception field is not limited. We can modify the definition of $\Omega_i$ to make the range small as the convolution filter or extended to the whole graph.

\begin{figure}
	\centering
	\includegraphics[width=0.45\textwidth]{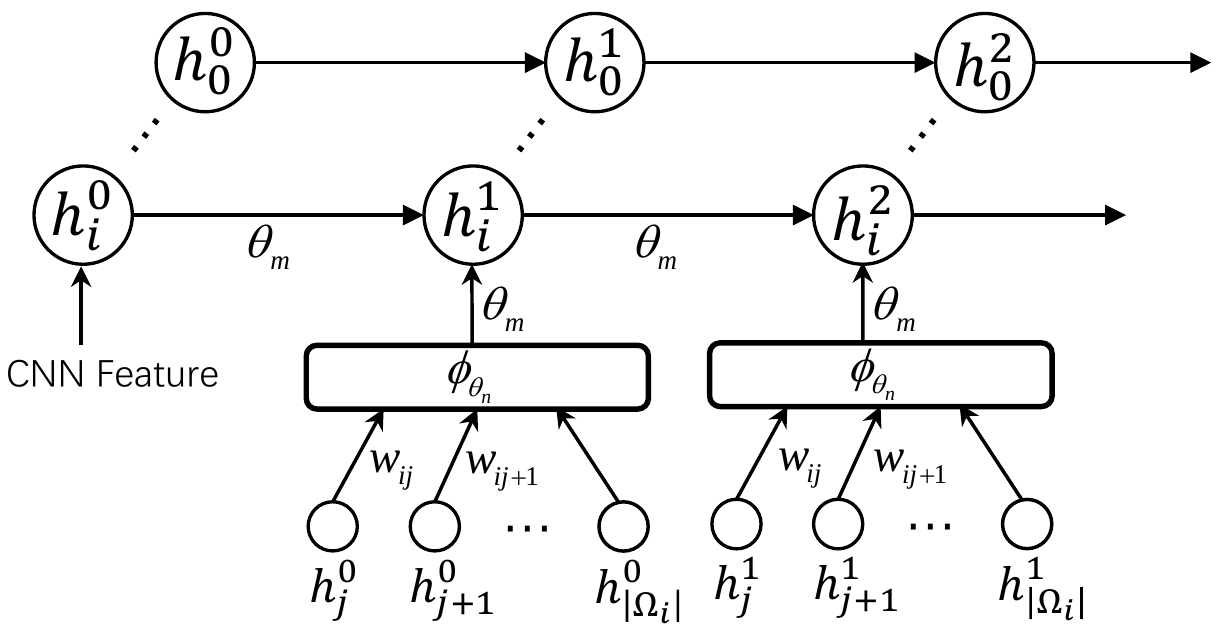}
	\caption{The unrolling of node updating. Nodes are initialized from CNN features, and updated via Eq.(\ref{lab_neighbor_msg}) and Eq.(\ref{lab_node_update}).}
	\label{fig:unroll_graph}
\end{figure}

\subsection{Node Updating}
The hidden state of node $v_i$ is denoted as $h_i$ which evolves with the graph iteration. The $h_i^t$ means the node state of $v_i$ at $t$ step. The initial value $h_t^0$ is obtained from the high dimensional output of ROI pooling as shown in Fig. \ref{fig:framework}. The node updating consists of two stages: firstly collecting neighbor message, then updating node state. As shown in Eq. (\ref{lab_neighbor_msg}), the function $\phi_{\theta_{n}}$ maps the neighbor nodes' messages at $t$ step into a vector $m_{i}^{t}$. The variable $w_{ij}$ decides the contributions of different nodes. $\phi_{\theta_{n}}$ is implemented with MLP in this paper.
\begin{equation}
\label{lab_neighbor_msg}
m_{i}^{t} = \frac{1}{|\Omega_i|}\sum_{j\in\Omega_i} \phi_{\theta_{n}}(w_{ij}h_{j}^{t}).
\end{equation}
After $m_{i}^{t}$ is determined, $\psi_{\theta_{m}}$ takes current node state $h_i^t$ and contextual message $m_{i}^{t}$ as input, and leads iteration to $t+1$ step obtaining $h_i^{t+1}$. This process is denoted as:
\begin{equation}
\label{lab_node_update}
h_{i}^{t+1} = Relu(\psi_{\theta_{m}}([h_{i}^{t}, m_{i}^{t}])), t\in [0,T] .
\end{equation}
$\phi_{\theta_{n}}$ and $\psi_{\theta_{m}}$ are shared by each node. The number of iterations $T$ is a manual designed hyper parameter. It also decides the range of contextual information, for example, increasing $T$ will make longer-range messages propagate to the target node.

\subsection{Node Prediction}
After Eq. (\ref{lab_node_update}) updating $T$ steps, we now predict the semantic label of each node. For a node $v_i$, the prediction $\hat{x}_i$ is obtained as following.   
\begin{equation}
\hat{x}_{i} = softmax(f_{\theta_{v}}(h_{i}^{T})), \hat{x}_i\in \mathbb{R}^{1\times{K}}.
\end{equation}
The final node state $h_{i}^{T}$ contains both its own state and neighbors' messages. $f_{\theta_{v}}$ is a MLP with the parameter $\theta_{v}$ which also is shared by all nodes. A softmax is concatenated after $f_{\theta_{v}}$. 

\subsection{Loss Function}
The cross entropy method is the most used loss function for this task. For one node $v_i$, it could not ensure the convergence of  $w_{ij}$ if we only consider the cross entropy of $v_i$. Furthermore, the center node loss and the neighbor node loss should be included at the same time, which is written as:
\begin{equation}
L = L_{center} + L_{neighbor}.
\end{equation}
Specifically, the $L_i$ for $v_i$ is defined as:
\begin{equation}
L_i = -(x_i ln(\hat{x}_i^T) + \frac{1}{|\Omega_i|}\sum_{j\in\Omega_i}(w_{ij}x_j ln(\hat{x}_j^T))),
\end{equation}
where $x_i$ is a one-hot vector denoted as the label of $v_i$. This equation consists of two items: the first item is the center loss, and the second item is neighbor loss. The function of neighbor loss is that it increases loss when the neighbor node associated with a high edge weight is wrongly predicted. Thus, the formal loss function is defined as following:
\begin{equation}
\label{eq:loss}
\begin{split}
L(X;\theta) &= \frac{1}{N}\sum_{i=1}^{N} L_i \\
&= -\frac{1}{N}\sum_{i=1}^{N}(x_i ln(\hat{x}_i^T) + \frac{1}{|\Omega_i|}\sum_{j\in\Omega_i}(w_{ij}x_j ln(\hat{x}_j^T))).
\end{split}
\end{equation}
The parameter $\theta=[\theta_{v}, \theta_{m}, \theta_{n}, \theta_{e}]$ and the CNN parameter $\Theta$ are joint learned with back-propagation algorithm during offline model training.

\begin{figure}
	\centering
	\includegraphics[width=0.45\textwidth]{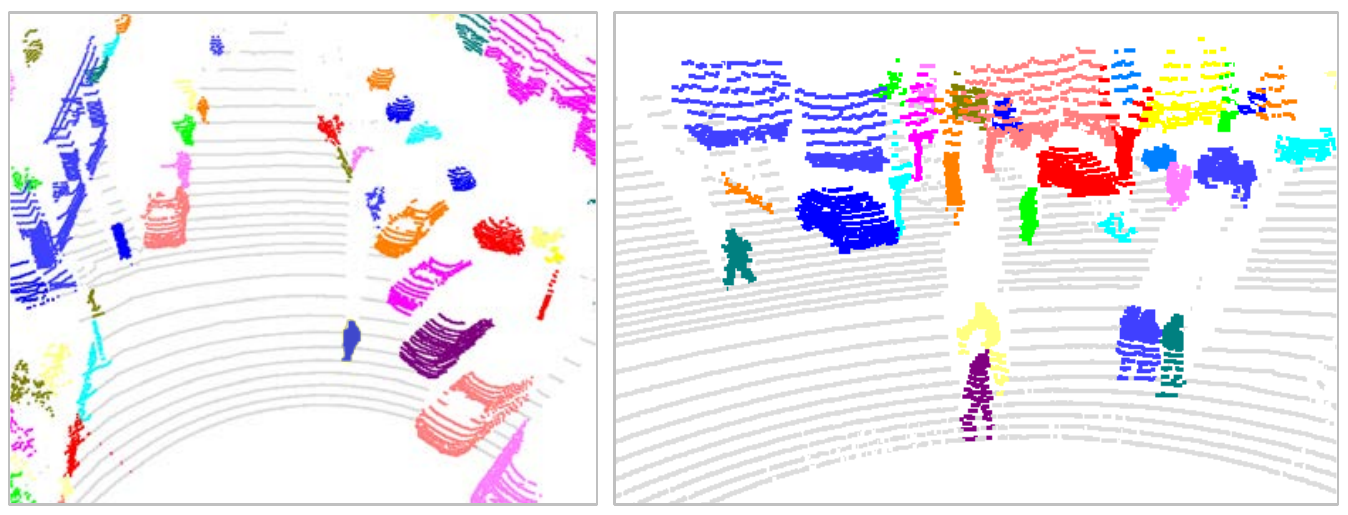}
	\caption{The results of segment generation. One color represents one segment.}
	\label{fig:segment}
\end{figure}

\section{Implementation Details}
\subsection{Segment Generation}
We use a 40-beams LiDAR sensor, and each scan line has 1800 points at different horizontal angles. That is, 40x1800 points are obtained for one data frame. These points are projected on a 2D matrix with the size (40,1800), and each row corresponds to a scan line. The pixel value of the matrix is the range measured by LiDAR sensor. Thus, the 2D matrix is named as range image. Note that the raw 3D point cloud and the range image could be converted into each other equally. The range image makes the unordered laser points grid-aligned, which benefits segment generation.

The segment generation is conducted on range image with a region grow based method. It is not reasonable applying a standard region grow method on range image, because the vertical and horizontal resolutions of range image are not equal, and these two parameters are decided by LiDAR sensor. On the other hand, large amounts of points hit on ground as the LiDAR sensor mounted on the top of platform. Removing ground firstly will enhance the effectiveness of segment generation. The result of segment generation is shown in Fig. \ref{fig:segment}.

Algorithm. \ref{segment_gen} gives the main process. In line 1, the $Background$ refers to the points whose distance to ground large than 4m. The $EdgePt$ is the target label of segment generation, because the edge points are the boundaries of region grow. The main function of line 2 is to remove ground. In line 3, an edge point is extracted by considering the distance to its 4-neighbors. More details of segment generation are given in the appendix of \cite{Mei2019IncorporatingHD}.

\begin{algorithm}
	\caption{\small Segment Generation}
	\label{segment_gen}
	\begin{algorithmic}[1]	
		\Require the point clouds $D$, and range image $A$
		\Ensure the segmentation result $B$ 
		\State \begin{varwidth}[t]{\linewidth}
			$L=\{Unvalid,Ground,Background,$ \par
			\hskip\algorithmicindent$Unknown,EdgePt\}$
		\end{varwidth}
		\State $S \gets CoarseSeg(D,L)$ \Comment{Removing ground.}
		\State $B \gets EdgeExtraction(A,S)$
		\State $B \gets RegionGrow(B)$
		\State return $B$
		
	\end{algorithmic}
\end{algorithm}

\subsection{Feature Network}
A fully convolution neural network is designed to extract feature for each node. The network is constructed as following: $C(3,64)+C(64,64)+M+C(64,128)+M+C(128,256)+D(256,256)+D(256,256)$, where C($m$, $n$) denotes a convolution layer whose input dimension is $m$ and output is $n$; $M$ means a maxpooling layer whose stride is 2;  $D(m,n)$ denotes a deconvolution layer whose stride is 2.  A batch-norm and relu layer  are concatenated for each $C$ and $D$. The width and height of the output feature map keep the same value with the input.

The input contains three raw feature channels. In Fig. \ref{fig:framework}, the bounding box is not ground truth,  but automatically generated from each segment on range image. Thus, one segment corresponds to one bounding box. And the bounding box is only used at the output of network. The raw features are range, intensity and height; the first two features are returned from LiDAR sensor. We linearly project range value from [0,25] meter to [0,255], if the range is large than 25 meters, it will be truncated at 255. As the ground is extracted, the height is defined as the vertical distance from the point to ground, and it is mapped from [0,5] meter to [0,255]. The intensity value is in [0,255]. 

A ROI pooling\cite{Ren2015FasterRT} layer is concatenated at the output of network. This layer crops ROIs on the feature map with specified bounding boxes. As aforementioned, one segment corresponds one box, thus, the hidden state of each segment is extracted from the output of CNN. Note that this layer unfolds each cropped feature into a fixed size vector, therefore, the hidden state of each segment shares the same feature size.

\subsection{Parameter learning}
The parameters in feature network, node updating and prediction are jointly learned via the stochastic gradient descent(SGD) method. The loss function is defined as Eq. (\ref{eq:loss}). We randomly choose one frame data at each iteration, so the batch size is 1. The initial learning rate is 0.01, and decays 0.5 at every 20 epochs. The total epoch number is 100 in our experiments.


\begin{figure}
	\centering
	\includegraphics[width=0.45\textwidth]{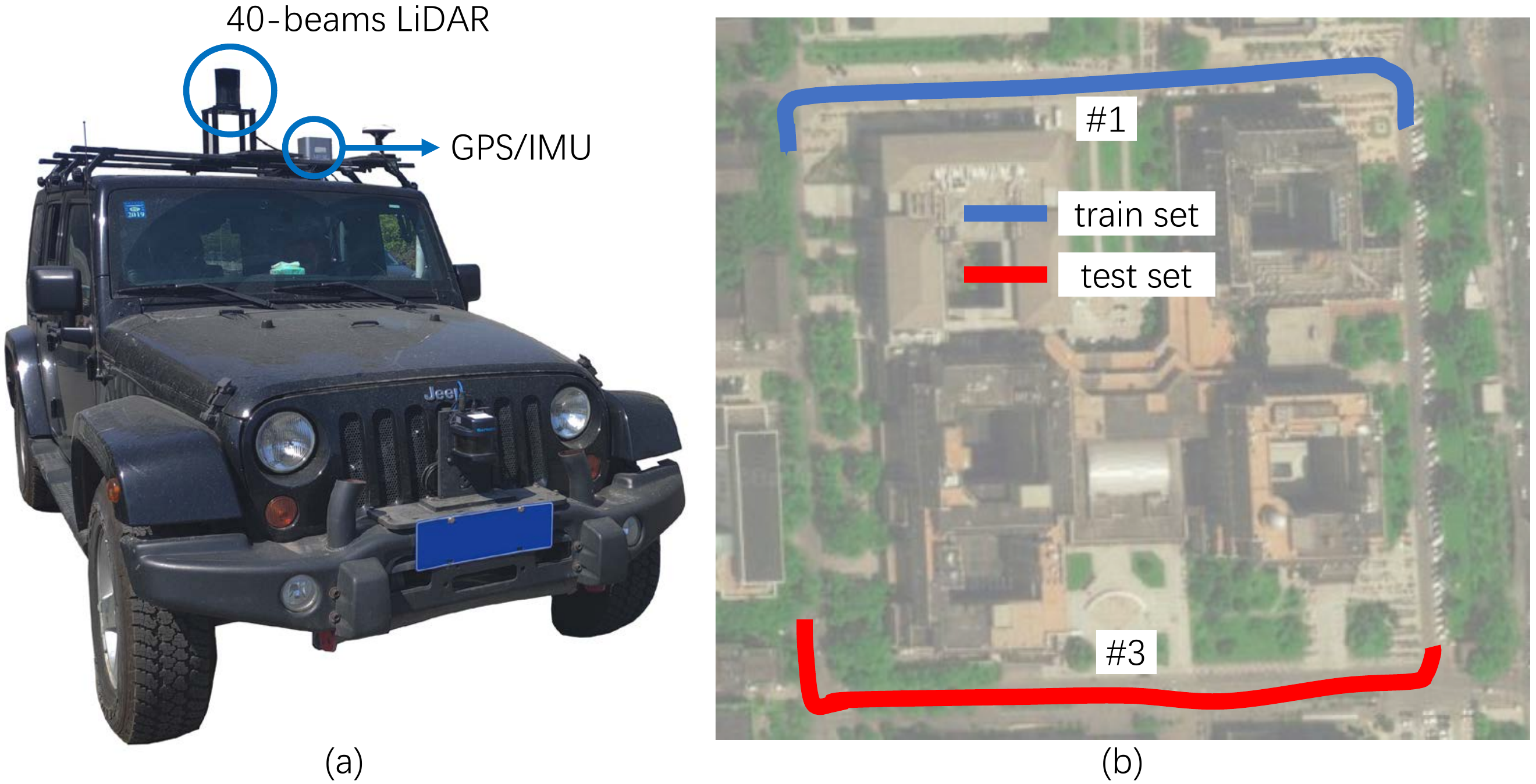}
	\caption{The data collection platform and the routes for experiments. The \#1 and \#3 of SemanticPOSS\cite{Pan2019SemanticPOSS} are used for train set and test set. }
	\label{fig:platform}
\end{figure}

\subsection{Graph Building}
In this part, we need define the node and determine the edge. The raw point cloud are clustered into multiple segments on the range image. One segment corresponds to one node. And the initial hidden state of the node is from the output of the ROI pooling layer. 

Here the edge only indicates the connectivity of target node and its neighborhoods. For the target node $v_i$, a hyper parameter $|\Omega_i|$ denoted as the neighbor number is set. Then, we obtain $|\Omega_i|$ neighbor nodes with top $|\Omega_i|$ shortest path to $v_i$. The path is defined as the center distance of two nodes. Finally, $v_i$ and its neighbor nodes are linked to build $|\Omega_i|$ edges.

\begin{figure*}
	\centering
	\includegraphics[width=1.0\textwidth]{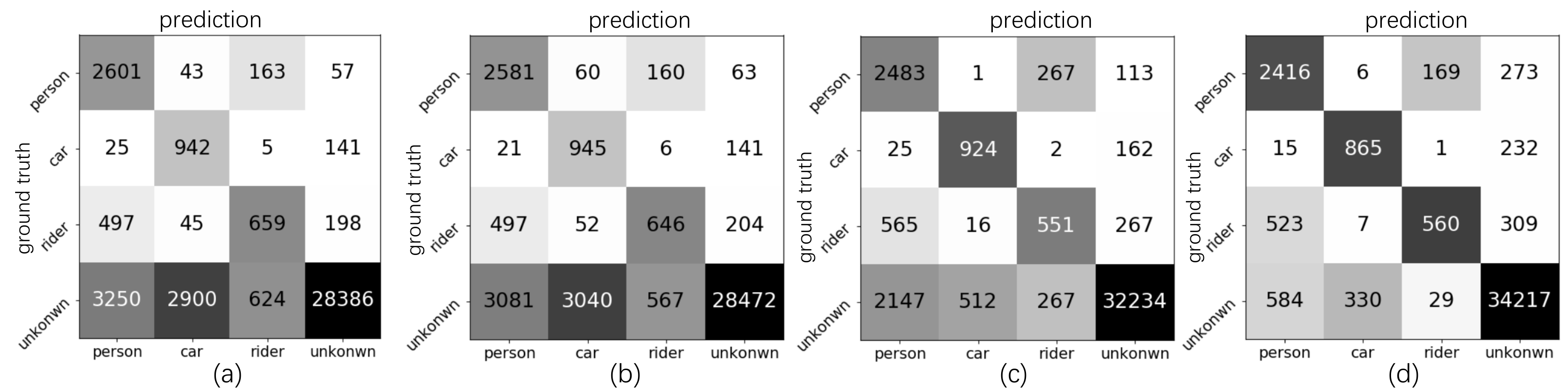}
	\caption{The confusion matrices for different classifiers. The color is indexed by precision, and darker color means higher precision. (a) Unary CNN. (b) 3DGNN-all. (c) 3DGNN-1nn. (d) The proposed one.  }
	\label{fig:cm}
\end{figure*}

\section{EXPERIMENTAL RESULTS}

\subsection{Dataset}

Our method is firstly evaluated on SemanticPOSS\cite{Pan2019SemanticPOSS} dataset. The collection platform is shown in Fig. \ref{fig:platform}, which equips a 40-beams LiDAR and GPS/IMU sensors. The total route is about 1.5 kilometers in the campus of Peking University, which is split into 6 sections. The LiDAR data is sequentially collected and annotated for every frame. The total frames are 2988. The difference with existing dataset, such as Semantic3D\cite{Hackel2017Semantic3DnetAN} and SemanticKITTI\cite{behley2019iccv}, lies in that SemanticPOSS contains large amounts of dynamic object, like car, people and rider. 

\begin{table}[]
	\centering
	\renewcommand\arraystretch{1.5}			
	\small
	
	\caption{The samples of manual annotations.}
	\label{tab:dataset}
	\begin{tabular}{|l|l|l|l|l|}
		\hline
		& People & Car  & Rider & Unknown \\ \hline
		train set & 4693   & 4888 & 911   & 14560   \\ \hline
		test set  & 2864   & 1113 & 1399  & 17379   \\ \hline
	\end{tabular}
\end{table}

As shown in Fig. \ref{fig:platform}(b), The section 1 and section 3 are used for train set and test set. Each section has 500 frames. Each frame is divided into multiple segments as illustrated in Fig. \ref{fig:segment}. The TABLE. \ref{tab:dataset} shows the number of segments, and actually we classify each segment. Note that a segment will not be included if the center range is larger than 25 meters or the center height is higher than 5 meters. Furthermore, we only consider dynamic objects like people, car and rider; and the static objects are categorized into unknown.

\subsection{Experimental Settings}

\begin{table*}[]
	\centering
	\renewcommand\arraystretch{1.5}			
	\small
	\caption{The quantitative results on test data.}
	\label{tab:rule-based}
	\begin{threeparttable}
		\begin{tabular}{|c|c|c|c|c|c|c|c|c|c|c|c|}
			\hline
			\multirow{2}{*}{} & \multicolumn{2}{c|}{People} & \multicolumn{2}{c|}{Car} & \multicolumn{2}{c|}{Rider} & \multicolumn{2}{c|}{Unkown} & \multicolumn{3}{c|}{average} \\ \cline{2-12} 
			& P     & R     & P     & R     & P     & R     & P     & R     & P              & R              & F1             \\ \hline
			Unary CNN     & 0.501 & 0.903 & 0.496 & 0.814 & 0.482 & 0.420 & 0.984 & 0.905 & 0.616          & \textbf{0.760} & 0.663          \\ \hline
			3DGNN-all & 0.418 & 0.901 & 0.231 & 0.849 & 0.468 & 0.462 & 0.810 & 0.810 & 0.482          & 0.755          & 0.572          \\ \hline
			3DGNN-1nn & 0.476 & 0.867 & 0.636 & 0.830 & 0.507 & 0.394 & 0.983 & 0.917 & 0.650          & 0.752          & 0.682          \\ \hline
			ours        & 0.683 & 0.844 & 0.716 & 0.777 & 0.738 & 0.400 & 0.977 & 0.973 & \textbf{0.778} & 0.750          & \textbf{0.749} \\ \hline
		\end{tabular}
		\begin{tablenotes}
			\footnotesize
			\item[1] P : Precision; R : Recall.
			\item[2] 3DGNN-* refer to \cite{Qi20173DGN}. 
		\end{tablenotes}
	\end{threeparttable}
\end{table*}

\begin{figure*}
	\centering
	\includegraphics[width=0.8\textwidth]{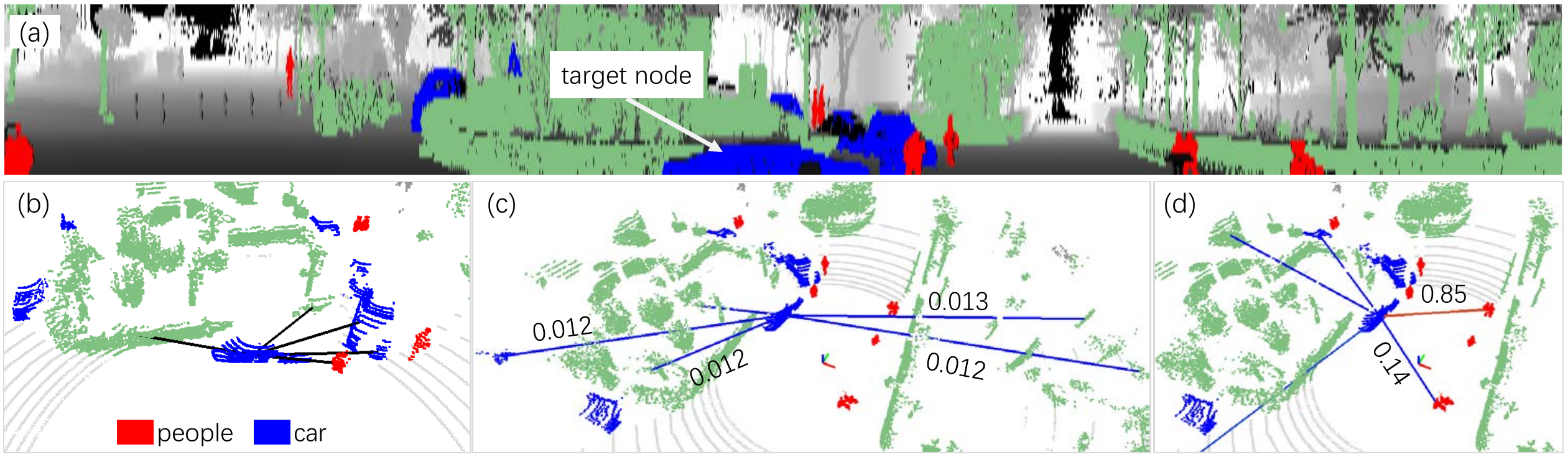}
	\caption{An illustration of edge weight updating. (a) A car is chosen as the target node on range image. (b) The edges of 3DGNN-5nn are fixed and each edge has the equal weight. (c) We sort the edge weights from high to low. The top-5 edges of our method at epoch-0 are depicted. Each edge has similar weight. (d) The top-5 edges of our method at epoch-15. The edge weights are updated automatically.}
	\label{fig:edge_updating}
\end{figure*}

Two baselines are compared with the proposed method. The first one is unary CNN, which does not include graph module and the output of ROI pooling in Fig. \ref{fig:framework} is directly sent to the prediction part. That is, the unary CNN is a conventional CNN-based classifier. The second one is denoted as 3DGNN\cite{Qi20173DGN}. We implement the same graph neural networks as \cite{Qi20173DGN}. As shown in Fig. \ref{fig:framework}, except not including edge weight estimation, 3DGNN shares the same configurations with the proposed method.

As mentioned in Sect. \ref{sect:method}.C, the contextual information are controlled by the number of neighbors. Thus, the neighbor number is a factor, for example, 3DGNN-1nn means taking 1 nearest neighbor, 3DGNN-5nn for 5 nearest neighbors, and so on. Note that 3DGNN-all means taking all neighbors. Our method takes all neighbor nodes into consideration, so we set the iteration number $T=1$.

The F1 measure is adopted for quantitative evaluations that is defined as below:
\begin{equation}
F1=\frac{2*Precision*Recall}{Precision+Recall}.
\end{equation}
The semantic segmentation is converted into classifying each node in this paper, so the F1 measure is reasonable.

\subsection{Compared with Unary CNN}
The details of different method are illustrated in TABLE. \ref{tab:rule-based} and Fig. \ref{fig:cm}. Our method has higher precision than unary CNN, which improves the mean F1 score by 8\%. Compared Fig. \ref{fig:cm}(a) and Fig. \ref{fig:cm}(d), the proposed one has better capability to discriminate objects from unknown.

\begin{figure*}
	\centering
	\includegraphics[width=1.0\textwidth]{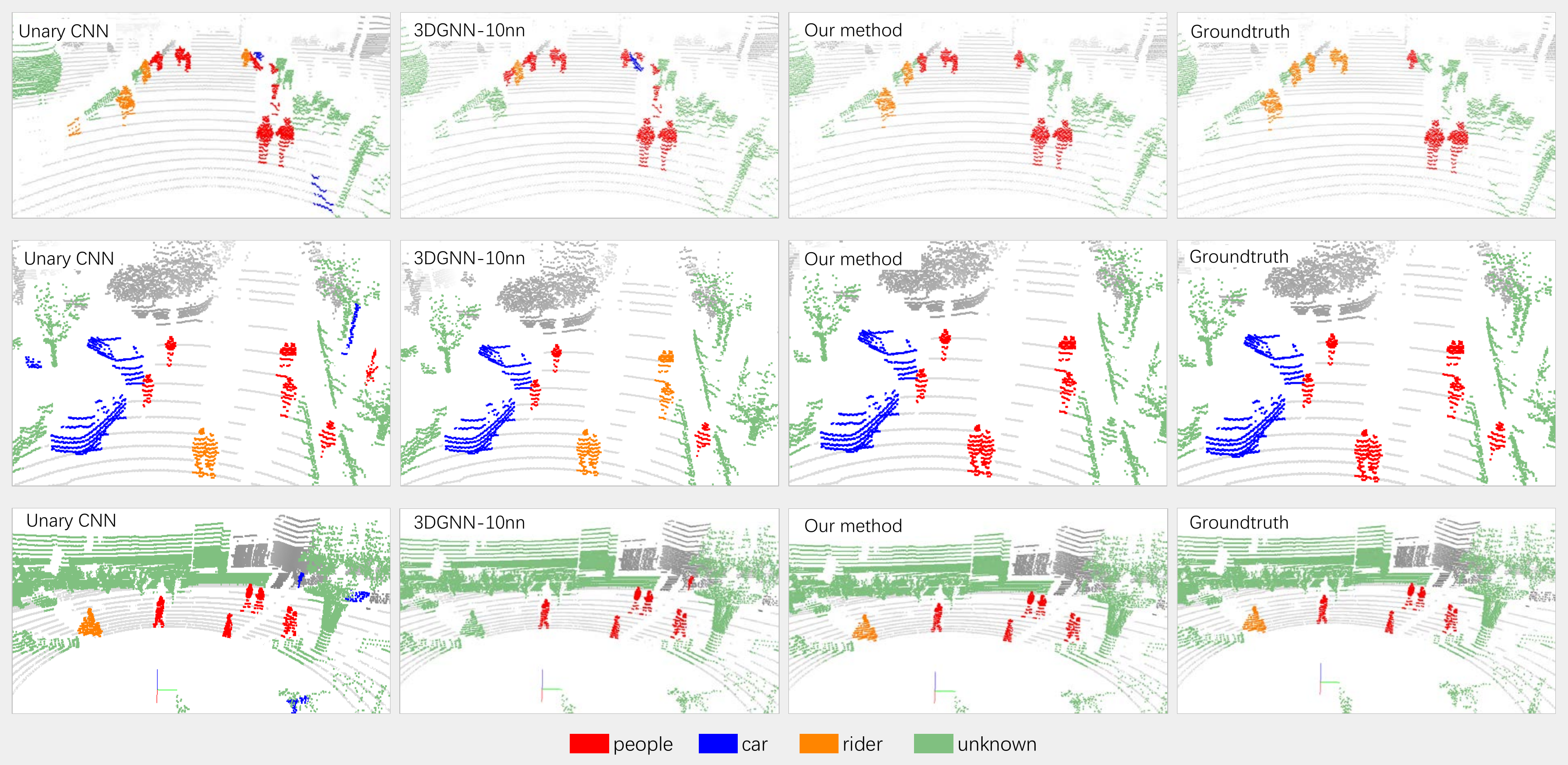}
	\caption{The qualitative comparisons of different methods.    }
	\label{fig:range_result}
\end{figure*}

The unary CNN does not consider the contextual information. One object like a car could be split into multiple nodes, therefore, unary CNN is more likely to classify these nodes into different categories as shown in the left column of  Fig.\ref{fig:range_result}. The conclusion that GNN enhances the unary CNN is also verified in \cite{Qi20173DGN}.

\subsection{Compared with 3DGNN} 
\begin{figure}
	\centering
	\includegraphics[width=0.5\textwidth]{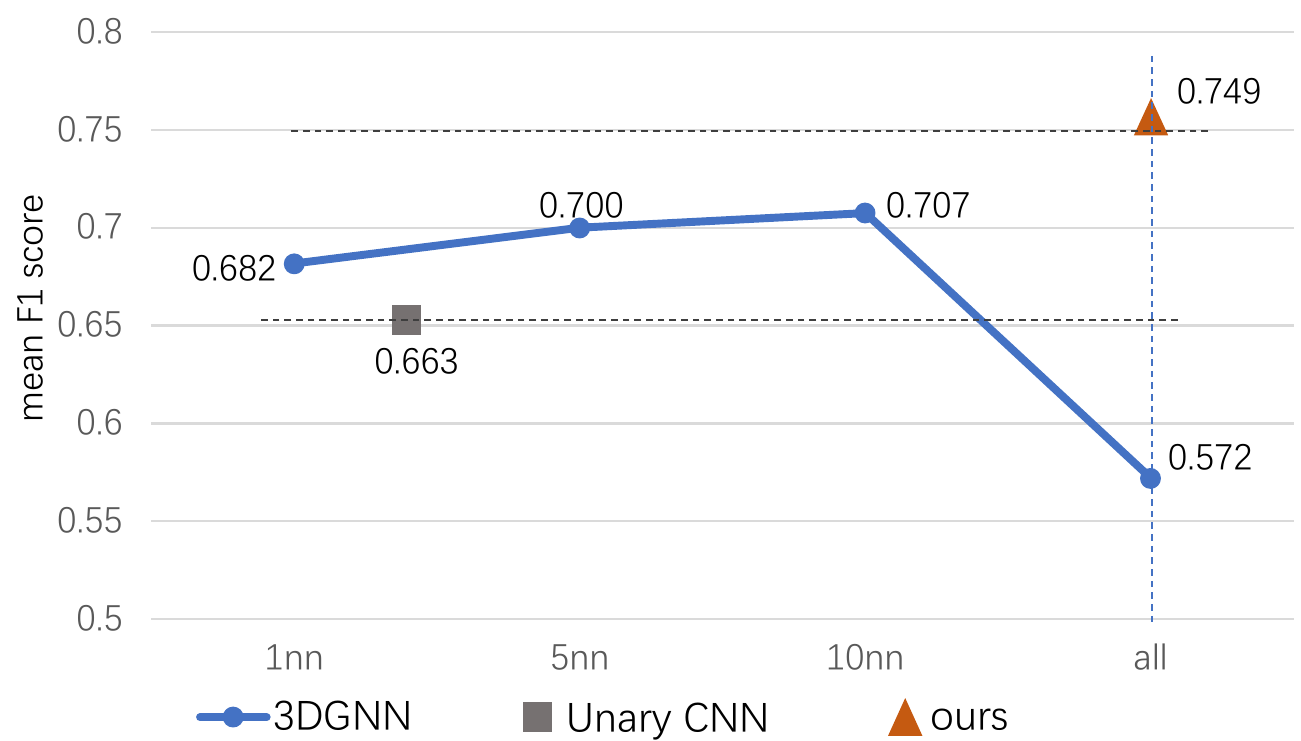}
	\caption{The comparisons with different methods. The horizontal axis tells how many neighbor nodes are adopted, for example, 5nn means 5 neighbor nodes. Note that Unary CNN does not considers neighbor relationships.  }
	\label{fig:f1curve}
\end{figure}

Here, we firstly make a comparison on considering all neighbors.  In Fig. \ref{fig:f1curve}, the F1 score of ours is higher than 3DGNN-all about 17\%. In TABLE. \ref{tab:rule-based}, these two methods have similar average recall, but the precision of ours is largely higher than 3DGNN. 

The difference of our method and 3DGNN lies in how to collect neighbor information. As shown in Eq. (\ref{lab_neighbor_msg}), the parameter $w_{ij}$ of 3DGNN is identical to $1$ for each neighbor; but $w_{ij}$ is automatically decided by a MLP in our method. Therefore, all neighbor nodes have the same contributions in 3DGNN; this strategy will hurt the performance of classifier when most of neighbor nodes are unrelated to the center node. However, our method has the ability to assign different weights for neighbor nodes avoiding the problem faced by 3DGNN.

\begin{figure}
	\centering
	\includegraphics[width=0.45\textwidth]{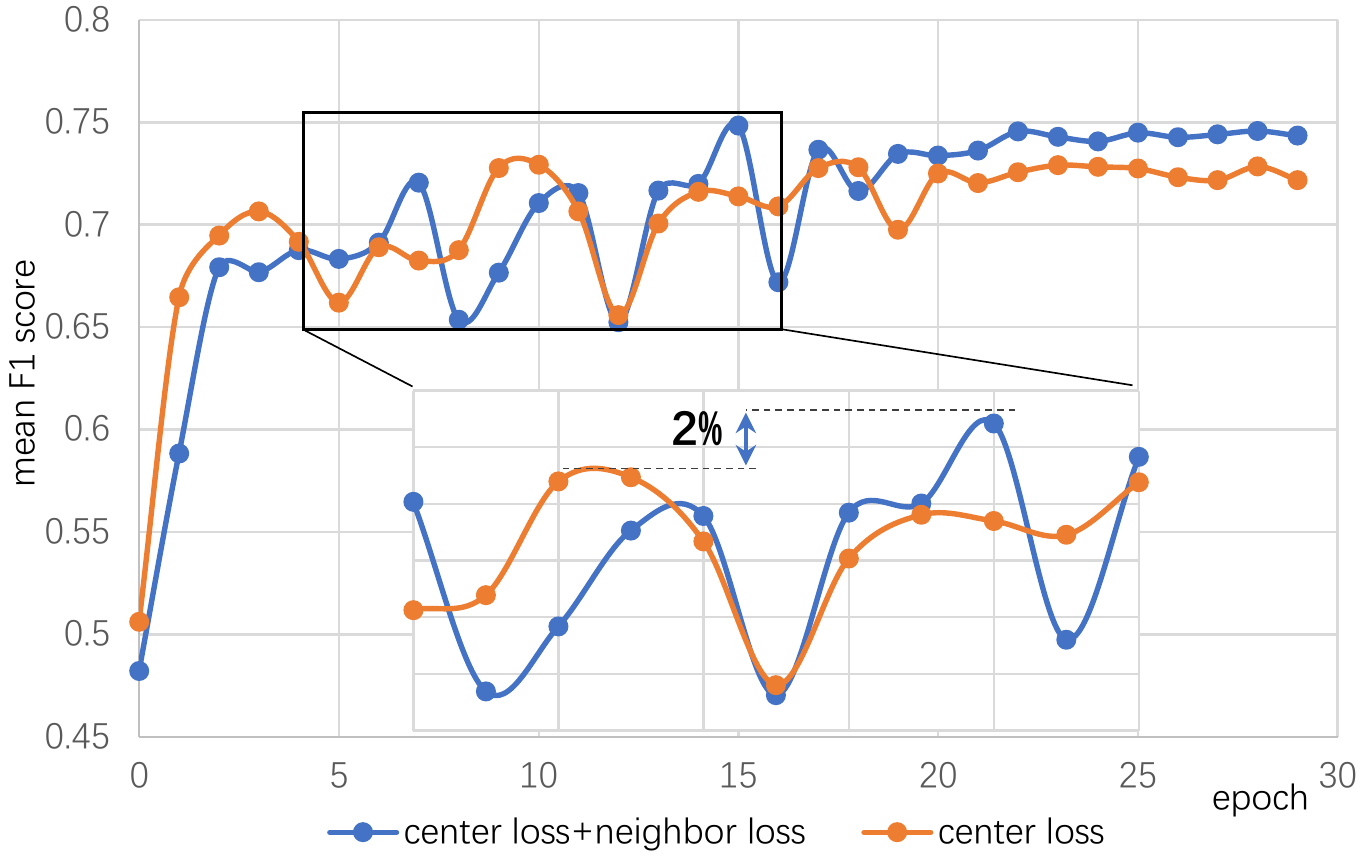}
	\caption{The comparisons with different loss functions on test set. The mean F1 scores are obtained by the evaluations on test set at each epoch.  The blue line is the loss function chosen by us. And the neighbor loss enhances the F1 score by 2\%.    }
	\label{fig:loss_compare}
\end{figure}

In Fig. \ref{fig:edge_updating}(d), the top-5 edge weights are shown, and we find that one edge weight usually is larger than others. That is, one neighbor dominates the contextual information. How does the 3DGNN behave if only one neighbor is considered?  Thus, the 3DGNN-1nn is listed in TABLE. \ref{tab:rule-based}. The average F1 score of 3DGNN-1nn is higher than 3DGNN-all about 11\%, which shows that the equal weight strategy is not suitable when considering large amount of neighbors. Then, the curve of 3DGNN with different neighbor configurations is obtained in Fig. \ref{fig:f1curve}.

In Fig. \ref{fig:f1curve}, different neighbor numbers make different F1 scores for 3DGNN. The neighbor number is a hyper parameter in previous works\cite{Qi20173DGN,Landrieu2018LargeScalePC}, and it will take many experiments finding an optimal parameter. However, the proposed method does not need the validation for this parameter, and demonstrates that it can still achieve high performance when considering all neighbor nodes.

\subsection{Validation for Loss Function}
The loss function is given in Eq. (\ref{eq:loss}). It contains center loss and neighbor loss. In Fig. \ref{fig:loss_compare}, the loss functions are validated on test set. We find that neighbor loss brings 2\% F1 score improvement than only considering center loss.

\section{conclusion and future work}
A scene context based semantic segmentation method is proposed for 3D LiDAR data. A GNN-based classifier is designed to capture contextual information. The problem is defined as building a graph to represent the topology of a center segment with its neighborhoods, then inferring the segment label. Evaluation on a dataset of dynamic scene indicates the effectiveness of the proposed method. Future work will address more experiments on other large-scale 3D LiDAR datasets. In addition, the introduce of new context, like temporal information, will also be studied.

\bibliographystyle{IEEEtran}

\begin{thebibliography}{10}
\providecommand{\url}[1]{#1}
\csname url@samestyle\endcsname
\providecommand{\newblock}{\relax}
\providecommand{\bibinfo}[2]{#2}
\providecommand{\BIBentrySTDinterwordspacing}{\spaceskip=0pt\relax}
\providecommand{\BIBentryALTinterwordstretchfactor}{4}
\providecommand{\BIBentryALTinterwordspacing}{\spaceskip=\fontdimen2\font plus
\BIBentryALTinterwordstretchfactor\fontdimen3\font minus
  \fontdimen4\font\relax}
\providecommand{\BIBforeignlanguage}[2]{{%
\expandafter\ifx\csname l@#1\endcsname\relax
\typeout{** WARNING: IEEEtran.bst: No hyphenation pattern has been}%
\typeout{** loaded for the language `#1'. Using the pattern for}%
\typeout{** the default language instead.}%
\else
\language=\csname l@#1\endcsname
\fi
#2}}
\providecommand{\BIBdecl}{\relax}
\BIBdecl

\bibitem{urmson2008autonomous}
C.~Urmson, J.~Anhalt, D.~Bagnell, C.~Baker, R.~Bittner, M.~Clark, J.~Dolan,
  D.~Duggins, T.~Galatali, C.~Geyer \emph{et~al.}, ``Autonomous driving in
  urban environments: Boss and the urban challenge,'' \emph{Journal of Field
  Robotics}, vol.~25, no.~8, pp. 425--466, 2008.

\bibitem{Shi2018PointRCNN3O}
S.~Shi, X.~Wang, and H.~Li, ``Pointrcnn: 3d object proposal generation and
  detection from point cloud,'' \emph{IEEE/CVF Conference on Computer Vision
  and Pattern Recognition (CVPR)}, pp. 770--779, 2018.

\bibitem{Mei2019SupervisedLF}
J.~Mei, J.~Chen, W.~Yao, X.~Zhao, and H.~Zhao, ``Supervised learning for
  semantic segmentation of 3d lidar data,'' \emph{IEEE Intelligent Vehicles
  Symposium (IV)}, pp. 1491--1498, 2019.

\bibitem{Zellers2017NeuralMS}
R.~Zellers, M.~Yatskar, S.~Thomson, and Y.~Choi, ``Neural motifs: Scene graph
  parsing with global context,'' \emph{IEEE/CVF Conference on Computer Vision
  and Pattern Recognition (CVPR)}, pp. 5831--5840, 2017.

\bibitem{Yang2018GraphRF}
J.~Yang, J.~Lu, S.~Lee, D.~Batra, and D.~Parikh, ``Graph r-cnn for scene graph
  generation,'' in \emph{European Conference on Computer Vision (ECCV)}, 2018.

\bibitem{Li2017SceneGG}
Y.~Li, W.~Ouyang, B.~Zhou, K.~Wang, and X.~Wang, ``Scene graph generation from
  objects, phrases and region captions,'' \emph{IEEE International Conference
  on Computer Vision (ICCV)}, pp. 1270--1279, 2017.

\bibitem{Zhou2018GraphNN}
J.~Zhou, G.~Cui, Z.~Zhang, C.~Yang, Z.~Liu, and M.~Sun, ``Graph neural
  networks: A review of methods and applications,'' \emph{ArXiv}, vol.
  abs/1812.08434, 2018.

\bibitem{Hackel2017Semantic3DnetAN}
T.~Hackel, N.~Savinov, L.~Ladicky, J.~D. Wegner, K.~Schindler, and
  M.~Pollefeys, ``Semantic3d.net: A new large-scale point cloud classification
  benchmark,'' \emph{ArXiv}, vol. abs/1704.03847, 2017.

\bibitem{behley2019iccv}
J.~Behley, M.~Garbade, A.~Milioto, J.~Quenzel, S.~Behnke, C.~Stachniss, and
  J.~Gall, ``{SemanticKITTI: A Dataset for Semantic Scene Understanding of
  LiDAR Sequences},'' in \emph{IEEE/CVF International Conference on Computer
  Vision (ICCV)}, 2019.

\bibitem{Qi20173DGN}
X.~Qi, R.~Liao, J.~Jia, S.~Fidler, and R.~Urtasun, ``3d graph neural networks
  for rgbd semantic segmentation,'' \emph{IEEE International Conference on
  Computer Vision (ICCV)}, pp. 5209--5218, 2017.

\bibitem{Wang2019GraphAC}
L.~Wang, Y.~Huang, Y.~Hou, S.~Zhang, and J.~Shan, ``Graph attention convolution
  for point cloud semantic segmentation,'' \emph{IEEE/CVF Conference on
  Computer Vision and Pattern Recognition (CVPR)}, pp. 10\,288--10\,297, 2019.

\bibitem{Liu2019PointVoxelCF}
Z.~Liu, H.~Tang, Y.~Lin, and S.~Han, ``Point-voxel cnn for efficient 3d deep
  learning,'' \emph{ArXiv}, vol. abs/1907.03739, 2019.

\bibitem{Landrieu2018LargeScalePC}
L.~Landrieu and M.~Simonovsky, ``Large-scale point cloud semantic segmentation
  with superpoint graphs,'' \emph{IEEE/CVF Conference on Computer Vision and
  Pattern Recognition (CVPR)}, pp. 4558--4567, 2018.

\bibitem{Pan2019SemanticPOSS}
Y.~Pan, B.~Gao, J.~Mei, S.~Geng, C.~Li, and H.~Zhao, ``Semanticposs: A point
  cloud dataset with large quantity of dynamic instances,'' \emph{ArXiv}, vol.
  abs/2002.09147, 2020.

\bibitem{munoz2009onboard}
D.~Munoz, N.~Vandapel, and M.~Hebert, ``Onboard contextual classification of
  3-d point clouds with learned high-order markov random fields,'' in
  \emph{IEEE International Conference on Robotics and Automation (ICRA)}.\hskip
  1em plus 0.5em minus 0.4em\relax IEEE, 2009, pp. 2009--2016.

\bibitem{zhao2010scene}
H.~Zhao, Y.~Liu, X.~Zhu, Y.~Zhao, and H.~Zha, ``Scene understanding in a large
  dynamic environment through a laser-based sensing,'' in \emph{IEEE
  International Conference on Robotics and Automation (ICRA)}.\hskip 1em plus
  0.5em minus 0.4em\relax IEEE, 2010, pp. 127--133.

\bibitem{lu2012simplified}
Y.~Lu and C.~Rasmussen, ``Simplified markov random fields for efficient
  semantic labeling of 3d point clouds,'' in \emph{IEEE/RSJ International
  Conference on Intelligent Robots and Systems (IROS)}.\hskip 1em plus 0.5em
  minus 0.4em\relax IEEE, 2012, pp. 2690--2697.

\bibitem{GarciaGarcia2017ARO}
A.~Garcia-Garcia, S.~Orts, S.~Oprea, V.~Villena-Martinez, and J.~A. Rodriguez,
  ``A review on deep learning techniques applied to semantic segmentation,''
  \emph{ArXiv}, vol. abs/1704.06857, 2017.

\bibitem{Caltagirone2017FastLR}
L.~Caltagirone, S.~Scheidegger, L.~Svensson, and M.~Wahde, ``Fast lidar-based
  road detection using fully convolutional neural networks,'' \emph{IEEE
  Intelligent Vehicles Symposium (IV)}, pp. 1019--1024, 2017.

\bibitem{hackel2017isprs}
T.~Hackel, N.~Savinov, L.~Ladicky, J.~D. Wegner, K.~Schindler, and
  M.~Pollefeys, ``{SEMANTIC3D.NET: A new large-scale point cloud classification
  benchmark},'' \emph{ISPRS Annals of the Photogrammetry, Remote Sensing and
  Spatial Information Sciences}, vol. IV-1-W1, pp. 91--98, 2017.

\bibitem{riegler2017octnet}
G.~Riegler, A.~O. Ulusoy, and A.~Geiger, ``Octnet: Learning deep 3d
  representations at high resolutions,'' in \emph{IEEE Conference on Computer
  Vision and Pattern Recognition (CVPR)}, vol.~3.\hskip 1em plus 0.5em minus
  0.4em\relax IEEE, 2017, pp. 6620--6629.

\bibitem{qi2017pointnet}
C.~R. Qi, H.~Su, K.~Mo, and L.~J. Guibas, ``Pointnet: Deep learning on point
  sets for 3d classification and segmentation,'' in \emph{IEEE Conference on
  Computer Vision and Pattern Recognition (CVPR)}.\hskip 1em plus 0.5em minus
  0.4em\relax IEEE, 2017, pp. 77--85.

\bibitem{Qi2017PointNetDH}
C.~R. Qi, L.~Yi, H.~Su, and L.~J. Guibas, ``Pointnet++: Deep hierarchical
  feature learning on point sets in a metric space,'' in \emph{NIPS}, 2017.

\bibitem{Velickovic2017GraphAN}
P.~Velickovic, G.~Cucurull, A.~Casanova, A.~Romero, P.~Li{\`o}, and Y.~Bengio,
  ``Graph attention networks,'' \emph{ArXiv}, vol. abs/1710.10903, 2017.

\bibitem{Mei2019IncorporatingHD}
J.~Mei and H.~Zhao, ``Incorporating human domain knowledge in 3d lidar-based
  semantic segmentation,'' \emph{ArXiv}, vol. abs/1905.09533, 2019.

\bibitem{Ren2015FasterRT}
S.~Ren, K.~He, R.~B. Girshick, and J.~Sun, ``Faster r-cnn: Towards real-time
  object detection with region proposal networks,'' \emph{IEEE Transactions on
  Pattern Analysis and Machine Intelligence}, vol.~39, pp. 1137--1149, 2015.

\end{thebibliography}

\end{document}